\documentclass[10pt, conference]{ieeeconf}

\IEEEoverridecommandlockouts 
\usepackage{textcomp}
\usepackage{gensymb}
\usepackage{blindtext}
\usepackage{graphicx}
\usepackage{amsmath}
\usepackage{amssymb}
\usepackage{algorithm}
\usepackage[usenames,dvipsnames,svgnames,table]{xcolor}
\usepackage[noend]{algpseudocode}
\usepackage[acronym]{glossaries} 
\usepackage[caption=false]{subfig}
\usepackage[all]{nowidow}
\usepackage{etoolbox}
\usepackage{threeparttable}
\usepackage{booktabs}
\usepackage{adjustbox}
\usepackage{optidef}
\usepackage{cellspace}
\makeatletter
\AfterEndEnvironment{algorithm}{\let\@algcomment\relax}
\AtEndEnvironment{algorithm}{\kern2pt\hrule\relax\vskip3pt\@algcomment}
\let\@algcomment\relax
\newcommand\algcomment[1]{\def\@algcomment{\footnotesize#1}}

\renewcommand\fs@ruled{\def\@fs@cfont{\bfseries}\let\@fs@capt\floatc@ruled
  \def\@fs@pre{\hrule height.8pt depth0pt \kern2pt}%
  \def\@fs@post{}%
  \def\@fs@mid{\kern2pt\hrule\kern2pt}%
  \let\@fs@iftopcapt\iftrue}
\makeatother

\newcommand\secref[1]{\mbox{(Sec. \ref{sec:#1})}}
\newcommand\figref[1]{\mbox{(Fig. \ref{fig:#1})}}
\newcommand\tfigref[1]{\mbox{Fig. \ref{fig:#1}}}

\newcommand\seclabel[1]{\label{sec:#1}}
\newcommand\figlabel[1]{\label{fig:#1}}

\newcommand\tbllabel[1]{\label{tbl:#1}}

\newcommand\citem[1]{\mbox{\cite{#1}}}

\newcommand\viewf{\mathbf{v}}
\newcommand\frontp{\mathbf{f}}
\newcommand\viewp{\mathbf{x}}
\newcommand\viewo{\boldsymbol{\phi}}
\newcommand\view{\mathbf{v}=\{\viewp,\viewo\}}
\newcommand\viewidx[1]{\mathbf{v}_#1=\{\viewp_#1,\viewo_#1\}}
\newcommand\fvpair{\mathbf{m}=(\viewf,\frontp)}
\newcommand\fvpairm{\mathbf{m}}

\newcommand\vray{\mathbf{w}}

\newcommand\point{\mathbf{p}}
\newcommand\cpoint{\mathbf{c}}
\newcommand\optpoint{\mathbf{s}}
\newcommand\npoint{\mathbf{n}}

\DeclareMathOperator*{\argmax}{arg\,max}
\DeclareMathOperator*{\argmin}{arg\,min}

\usepackage[utf8]{inputenc}
\DeclareUnicodeCharacter{0229}{e}
\usepackage[
 bibstyle=ieee,
 citestyle=numeric-comp,
 isbn=false,
 doi=false,
 eprint=false,
 sorting=none,
 url=true,
 bibencoding=utf8,
 backend=biber
]{biblatex}
\DeclareFieldFormat{url}{\url{#1}}
\DeclareBibliographyCategory{needsurl}
\newcommand{\entryneedsurl}[1]{\addtocategory{needsurl}{#1}}
\renewbibmacro*{url+urldate}{%
  \ifcategory{needsurl}
    {\printfield{url}%
     \iffieldundef{urlyear}
       {}
       {\setunit*{\addspace}%
        \printurldate}}
    {}}

\title{Proactive Estimation of Occlusions and Scene Coverage for\\ Planning Next Best Views in an Unstructured Representation}

\author{Rowan Border$\,^{1}$ and Jonathan D. Gammell$\,^{1}$
\thanks{$^{1}\,$Rowan Border and Jonathan D. Gammell are with the Estimation, Search, and Planning (ESP) Research Group, Oxford Robotics Institute (ORI), Department of Engineering Science, University of Oxford, Oxford, United Kingdom. {\tt\small \{rborder,gammell\}@robots.ox.ac.uk}}
}

\begin{document}

\entryneedsurl{radcliffe}

\maketitle

\IEEEpeerreviewmaketitle

\newacronym{see}{SEE}{Surface Edge Explorer}
\newacronym{nbv}{NBV}{Next Best View}

\begin{abstract}

The process of planning views to observe a scene is known as the \gls{nbv} problem. Approaches often aim to obtain high-quality scene observations while reducing the number of views, travel distance and computational cost.

Considering occlusions and scene coverage can significantly reduce the number of views and travel distance required to obtain an observation. Structured representations (e.g., a voxel grid or surface mesh) typically use raycasting to evaluate the visibility of represented structures but this is often computationally expensive. Unstructured representations (e.g., point density) avoid the computational overhead of maintaining and raycasting a structure imposed on the scene but as a result do not proactively predict the success of future measurements.

This paper presents proactive solutions for handling occlusions and considering scene coverage with an unstructured representation. Their performance is evaluated by extending the density-based \gls{see}. Experiments show that these techniques allow an unstructured representation to observe scenes with fewer views and shorter distances while retaining high observation quality and low computational cost.

\end{abstract}

\glsresetall

\section{Introduction}

High-quality 3D observations of the real world are valuable for performing infrastructure analysis and creating realistic simulations. A bounded region of space containing structures (i.e., a \emph{scene}) is often observed with a depth sensor that is actuated by a robotic or human-operated platform.      

Finding a set of views to observe a scene is known as the \gls{nbv} planning problem. Approaches to the \gls{nbv} problem seek to obtain high-quality scene observations while reducing the number of views taken, the distance travelled and/or the associated computational cost. 

Scenes can be observed using fewer views and less travelling by planning views with good visibility of incompletely observed surfaces that are close to the sensor position. This is typically achieved by accounting for occlusions and scene coverage when proposing and selecting next best views.

\begin{figure}[tpb]
	\centering
	\captionsetup[subfigure]{labelformat=empty}
	\subfloat[]{\includegraphics[width=\linewidth]{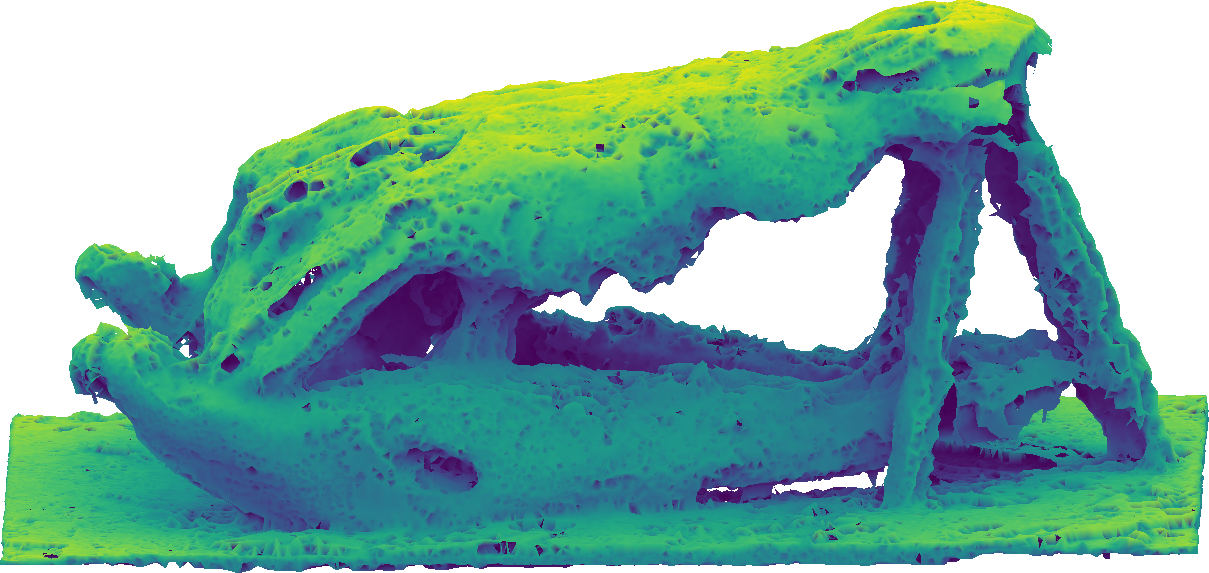}}
	\caption{A mesh reconstruction of the pointcloud obtained by SEE++ from observing a saltwater crocodile (\emph{Crocodylus porosus}) skull \citem{Schneider1801} with a hand-held sensor. Considering occlusions and scene coverage reduced the number of views by $41\%$ ($134$ vs. $226$) and distance travelled by $46\%$ ($46\,$m vs. $85\,$m) at the cost of an $187\%$ increase in computation time ($158\,$s vs. $55\,$s).}
	\figlabel{marquee}
	\vspace{-2ex}
\end{figure}
 
Structured representations can detect occlusions and evaluate surface coverage by raycasting their imposed structure (i.e., mesh triangles for surface representations or voxels for volumetric structures). Raycasting provides valuable knowledge for selecting views but it is often computationally expensive and does not aid the proposal of unoccluded views.   

Unstructured representations (e.g., point density) alternatively reason directly about measurements. Their use of point-based scene knowledge mitigates the computational cost of maintaining an imposed structure and does not constrain the fidelity of represented information; however, point representations preclude the use of traditional methods (e.g., raycasting) for evaluating occlusions and scene coverage.

This paper presents strategies for proactively handling occlusions and considering scene coverage with an unstructured representation. These techniques detect point-based occlusions, optimise unoccluded view proposals and evaluate pointwise visibility to select next best views that most improve an observation while travelling short distances. 

The presented methods reduce the number of views and travel distance required to obtain complete observations. This is shown by extending the \gls{see} \citem{Border2018}, a \gls{nbv} approach with an unstructured density representation. 

SEE++ is compared experimentally to SEE and state-of-the-art volumetric approaches \citem{Vasquez-Gomez2015, Kriegel2015, Delmerico2017} in simulation. Three standard models \citem{Newell1975, Turk1994, Curless1996} and a full-scale building model \citem{radcliffe} are observed with simulated depth sensors. The results show that SEE++ consistently requires fewer views and shorter travel distances than the other evaluated approaches to obtain an equivalent quality of scene observations.

Real world results for SEE and SEE++ are demonstrated with the observation of a saltwater crocodile (\emph{Crocodylus porosus}) skull \citem{Schneider1801} using an Intel Realsense D435 \figref{marquee}.  

This paper is organised as follows. Section II presents a review of existing methods to account for occlusions and scene coverage. Section III presents point-based approaches for proactively handling occlusions and scene coverage. Section IV presents both statistically significant comparisons of \gls{nbv} approaches in a simulated environment and a real-world demonstration of \gls{see} and SEE++. Sections V and VI discuss the results and plans for future work. 

\section{Related Work}

Scott et al. \citem{Scott2003a} present a two-dimensional categorisation of \gls{nbv} approaches. Techniques are classified by their scene representation and whether the approach is model-free or model-based. Model-based approaches require an \emph{a priori} scene model to plan a view path. These approaches are suited for inspection tasks that compare real-world models with a known ground truth but do not generalise to unknown scenes.

Model-free approaches plan next best views based on previous measurements and do not require \emph{a priori} scene models. These approaches commonly represent the scene with a volumetric or surface representation. A volumetric representation discretises the scene into a 3D voxel grid that represents whether volumes of space contain measurements. Surface representations approximate the scene geometry by connecting measurements into a triangulated surface mesh. 

\subsection{Volumetric Approaches}

Methods for proposing and selecting next best views are often tied to the representation used. Volumetric-based approaches \citem{Monica2018, Bircher2018a, Song2017, Maniatis2017, Daudelin2017, Delmerico2017, Potthast2014, Vasquez-Gomez2015, Curless2011, Connolly1985} commonly evaluate scene visibility by raycasting their voxel grid from proposed views to determine which voxels are observable. These algorithms typically quantify view quality based on the number of visible voxels and measurement density within each voxel. 

The view proposal problem is frequently simplified in volumetric-based approaches by initialising a fixed set of views surrounding the scene \citem{Maniatis2017, Daudelin2017, Delmerico2017, Potthast2014, Vasquez-Gomez2015, Curless2011, Connolly1985}. This removes the complexity of proposing views based on sensor measurements but prevents approaches from adjusting views to account for occlusions and scene coverage. As a result, the observation quality obtained with these approaches is highly dependent on the density and distribution of the fixed views.

Some volumetric approaches obtain high-quality scene models by proposing views using path planning algorithms. Bircher et al. \citem{Bircher2018a} use an RRT \citem{LaValle1998} to grow an exploration tree from the current sensor position through empty voxels in the scene. Tree generation is stopped when a given number of nodes are created or a node is found with a non-zero number of visible unobserved voxels. The next best view is the node from which the greatest number of unobserved voxels are visible. Selin et al. \citem{Selin2019} improve upon this approach by computing the continuous (i.e., not voxel aligned) volume of unobserved space visible from proposed views using cubature integration and a novel sparse raycasting technique. 

Song et al. \citem{Song2017} present a similar approach using RRT* \citem{Karaman2011} but also consider the number of unobserved voxels visible from the path between the current sensor position and the potential next best view. They identify a minimal set of intermediate views sufficient to observe all of the unobserved voxels visible from the path and increase the completeness of the scene model obtained. This approach is extended in \citem{Song2018} with a surface representation that adapts views to account for occlusions, and as a result improves the visibility of surfaces.

Occlusions are not actively addressed by these approaches, except for in \citem{Song2018}, as they evaluate the observability of scene volumes but do not adapt views to improve their visibility.  

\subsection{Surface Approaches}

Techniques using a surface representation \citem{Dierenbach2016, Khalfaoui2013, Roberts2017, Karaszewski2016, Cunningham-Nelson2015, Hollinger2012, Trummer2010} can identify occlusions by raycasting their triangulated surface mesh. Surface coverage is improved by proposing views orthogonal to the mesh surface at detected boundaries. These boundaries can either be outer edges of the mesh or holes resulting from insufficient measurements. Many surface-based approaches use a multistage observation process that first obtains an initial surface mesh from preplanned views and then proposes additional views to improve it \citem{Roberts2017, Karaszewski2016, Cunningham-Nelson2015, Hollinger2012, Trummer2010}.

Dierenbach et al. \citem{Dierenbach2016} and Khalfaoui et al. \citem{Khalfaoui2013} present approaches that do not require multistage observations. Dierenbach et al. \citem{Dierenbach2016} use the Growing Neural Gas algorithm \cite{Fritzke1995} to incrementally construct a surface mesh from point measurements. A 3D Voronoi tessellation is then computed and the mesh vertex in the Voronoi cell with the lowest density is selected as the target for the next best view. The view is placed at a distance along the surface normal defined as a function of the sensor resolution and scene size. This approach is shown to obtain high-quality models but some surfaces may be unobserved as occlusions are not considered.

Khalfaoui et al. \citem{Khalfaoui2013} obtain high-quality scene observations by accounting for surface occlusions and selecting views that improve scene coverage. Each point measurement in the triangulated mesh is classified as either fully or partially visible based on the angle between the local surface normal and the poses of previous views. If a point is occluded from all previous views then its surface normal, as defined by the mesh, is added to the set of view proposals. These view proposals are clustered using the mean shift algorithm and the closest cluster center is selected as the next best view. 

\subsection{Other Approaches}

Kriegel et al. \citem{Kriegel2015} use a combined surface and volumetric representation. Views are proposed to observe the boundaries of a triangulated surface mesh. Next best views are selected by considering the surface quality of the triangulated mesh representation and the observation states of voxels in the volumetric representation. Occlusions are handled by rotating the view relative to the target surface until it can be observed. 

\gls{see} \citem{Border2018} uses a density representation. Measurements are classified based on the number of neighbouring points within a given radius and views are proposed to observe surfaces with insufficient measurements. Occlusions are handled reactively by capturing incrementally adjusted views until the target surface is successfully observed. Next best views are selected to be close to the sensor position but the coverage of scene surfaces from proposed views is not considered.

\vspace{1ex}

This paper presents point-based methods to proactively handle occlusions and consider the coverage of scene surfaces when planning next best views with an unstructured representation. These techniques detect occlusions, optimise views to avoid occluding points and select views that most improve surface coverage while travelling short distances. The resulting reduction in the travel distance and views required to observe a scene is demonstrated with SEE++.

\section{SEE++}
\seclabel{overview}

\gls{nbv} planning approaches can typically observe scenes more efficiently by considering occlusions and scene coverage when proposing and selecting next best views. The methods presented in this paper allow approaches with unstructured scene representations to proactively consider point-based occlusions and scene coverage. The advantages of these techniques are demonstrated with SEE++, an extension of \gls{see} \citem{Border2018} that uses an unstructured density representation. 

\gls{see} aims to observe scenes with a minimum desired measurement (i.e., point) density, $\rho$, by evaluating the number of points within a given resolution radius, $r$, of each sensor measurement. The desired density is chosen to attain the structural detail required for a given application (e.g., infrastructure inspection). The resolution radius should be sufficiently large to robustly handle measurement noise without incurring a significant increase in computational cost.  

Points with a sufficient density of neighbouring measurements are classified as \emph{core} and those without are classified as \emph{outliers}. The boundary between completely and partially observed scene regions is identified by classifying outlier points that have both core and outlier neighbours as \emph{frontiers}.

Scene coverage is expanded by obtaining new measurements around these frontier points. Views are proposed by estimating the local surface geometry around frontiers and placing a view at a given distance, $d$, along the surface normal of each frontier. A next best view is selected from this set of view proposals to reduce the sensor travel distance. Occlusions are addressed reactively by applying incremental view adjustments when a target frontier point is not observed. Views are selected until there are no more frontier points.

This paper presents techniques for proactively handling known occlusions when proposing views. Accounting for occluding points before attempting to observe a frontier reduces the number of views and sensor travel distance required to observe a scene. This is the result of requiring fewer incremental view adjustments to observe frontier points as known occlusions are avoided before views are obtained. 

A frontier point is considered occluded from a view if there are point measurements within an $r$-radius of the proposed sight line from the view position to the frontier point \secref{occ_detect}. This is used to detect occlusions for the $\tau$-nearest view proposals to the current sensor position. Occluded view proposals are updated to avoid known occlusions by considering the occluding points within a given occlusion search distance, $\psi$, of each frontier point and finding the furthest sight line from any potential occlusion \secref{view_opt}.

Detecting occlusions also makes it possible to consider surface coverage when selecting a next best view \secref{view_sel}. The visibility of frontier points from different views is captured with a directed graph. This \emph{frontier visibility graph} connects each frontier to the views from which it can be observed. The next best view is chosen from the graph to have the greatest number of outgoing edges (i.e., visible frontiers) relative to the distance from the current sensor position. This constrains the sensor travel distance while providing high coverage of incompletely observed surfaces. 

\subsection{Detecting Occlusions}
\seclabel{occ_detect}

\begin{figure}[tpb]
 \centering
  \includegraphics[width=\linewidth]{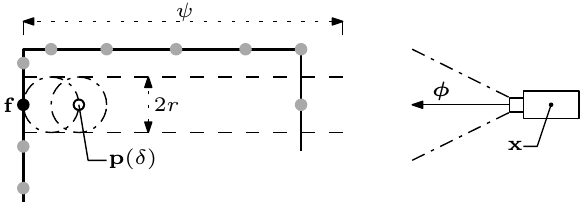}
  \caption{A cross sectional illustration of the occlusion detection approach. Points (grey dots) that would occlude the visibility of a frontier point (black dot), $\frontp$, from a proposed view are found by searching within the resolution radius, $r$, of points, $\point(\delta)$, along the sight line between the view position, $\viewp$, and the frontier point up to a given occlusion search distance, $\psi$. If the view was proposed to observe the frontier then the vector representing the sight line, $\vray$, is equivalent to the view orientation, $\vray = \viewo$.}
  \figlabel{occ_detect}
  \vspace{-2ex}
\end{figure}

A frontier point can only be successfully observed if sufficient measurements are obtained within its $r$-radius to reclassify it as a core point. This requires the sight line between a view and the frontier point to be free of occlusions. Occluding points are detected by searching for measurements within an $r$-radius of the sight line. A view is considered occluded if any points are found \figref{occ_detect}.      

A view, $\view$, is defined by a position, $\viewp$, and an orientation, $\viewo$. The sight line, $\vray$, between a frontier point, $\frontp_j$, and a given view, $\viewf_i$, is defined as the normalised vector from the frontier point to the view position, 
\[\vray_{ji} = \frac{\viewp_i - \frontp_j}{||\viewp_i - \frontp_j||} \,.\] 

Occlusions are found by searching for measurements within an $r$-radius of points along the sight line at an interval equal to the resolution radius. The set of occluding points between the view and frontier point is the union of the sets of neighbouring points within the search radius of each point, $\point(\delta) = \frontp_j + \delta\vray_{ji}$,
\[O(\frontp_j, \viewf_i) := \bigcup\limits_\delta N(P, r, \point(\delta))\,,\quad \delta = \zeta\,,\zeta+r\,,\dots,\psi \,,\]
where $\psi$ is the occlusion search distance, $\zeta$ is an offset along the sight line, $P$ is the set of observed points and $N(P, r, \point(\delta))$ is the set of points within an $r$-radius of the point $\point(\delta)$, e.g.,
\[N(P,r,\point) := \{\mathbf{q} \in P \;|\;||\mathbf{q} - \point|| \leq r\}\,.\]
An empty set of occluding points denotes that the frontier point is visible from the proposed view.

Checking the entire sight line for occlusions is computationally expensive and sensitive to surface noise. In practice, it is sufficient to detect occlusions up to a given occlusion search distance, $\psi$, from the frontier starting at an offset, $\zeta$, along the sight line. This search distance is chosen based on the view distance and structural complexity of the scene.   

A suitable offset, $\zeta$, for a frontier point is determined by considering points along the sight line of its observing view, $\viewf_\mathrm{o}$. Points that exist within an $r$-radius of the sight line between the observing view and the frontier could have occluded its visibility but evidently did not as the frontier point was observed. Their presence indicates that points closer to the frontier than this distance are unlikely to obstruct visibility. This offset is found by performing an occlusion search, as described above, along the sight line of the observing view until a point is reached with no neighbouring measurements.

The occlusions detected with this approach inform the proposal of unoccluded views and the connectivity of the frontier visibility graph used for selecting next best views. 

\subsection{Proactively Handling Occlusions}
\seclabel{view_opt}

A new unoccluded view is proposed for a frontier point when its current proposed view is classified as occluded. A suitable view is found by maximising the separation between a potential sight line and any view direction from which the frontier point is known to be occluded. This ensures that the clearest view of the frontier is proposed given the current knowledge of potential occlusions \figref{view_opt}.

The view directions from which the frontier point is occluded are denoted by the relative orientations of occluding points within the occlusion search radius. These directions can be represented as points on a sphere by normalising the distance of occluding points from the frontier. The maximal separation of a sight line from the occluded view directions is found by maximising the minimum distance between the normalised points and an optimised point on the sphere (i.e., a \emph{maximin} optimisation on a sphere). 

The use of a spherical projection to preserve the relative orientation of occluding points while normalising their distance is inspired by Hidden Point Removal \citem{Katz2007}. Points within the occlusion search distance of the frontier, $\frontp$, are projected onto a unit sphere around a central point, $\cpoint$, 
\[Q = \left\lbrace \frac{\point-\cpoint}{||\point-\cpoint||} \,\middle|\, \point \in N(P,\psi, \frontp) \right\rbrace\,.\]
The maximin solution is a point on this unit sphere which maximises the minimum distance to the projected points, $Q$.  

In an idealised scenario with no sensor noise the projection center is the frontier point. This ensures occluded view directions are accurately represented on the unit sphere. In practice it is necessary to offset the projection center from the frontier to prevent nearby points that are unlikely to occlude visibility from blocking valid view directions. The occlusion detection offset is reused as the projection center as it represents a point approximately clear of surface noise.  

A solution to the maximin optimisation on a sphere is the antipole of a solution to the \emph{minimax} problem \citem{Wesolowsky1983} (i.e., a point on the sphere which minimises the maximum distance to the projected points). The minimax solution is the center of the smallest spherical cap containing all of the projected points \citem{Patel2002}. This cap is defined by the pose of a plane intersecting the sphere. The solution is found by optimising the orientation of the plane normal, $\npoint$, and its distance from the center of the sphere, $e$. 

The plane normal points towards the smaller of the two spherical caps defined by the plane intersection. It is initialised using the orientation of the view from which the frontier point was first observed, $\viewo_\mathrm{o}$, as this sight line is known to be unoccluded. 

\begin{figure}[tpb]
 \centering
  \includegraphics[width=0.5\linewidth]{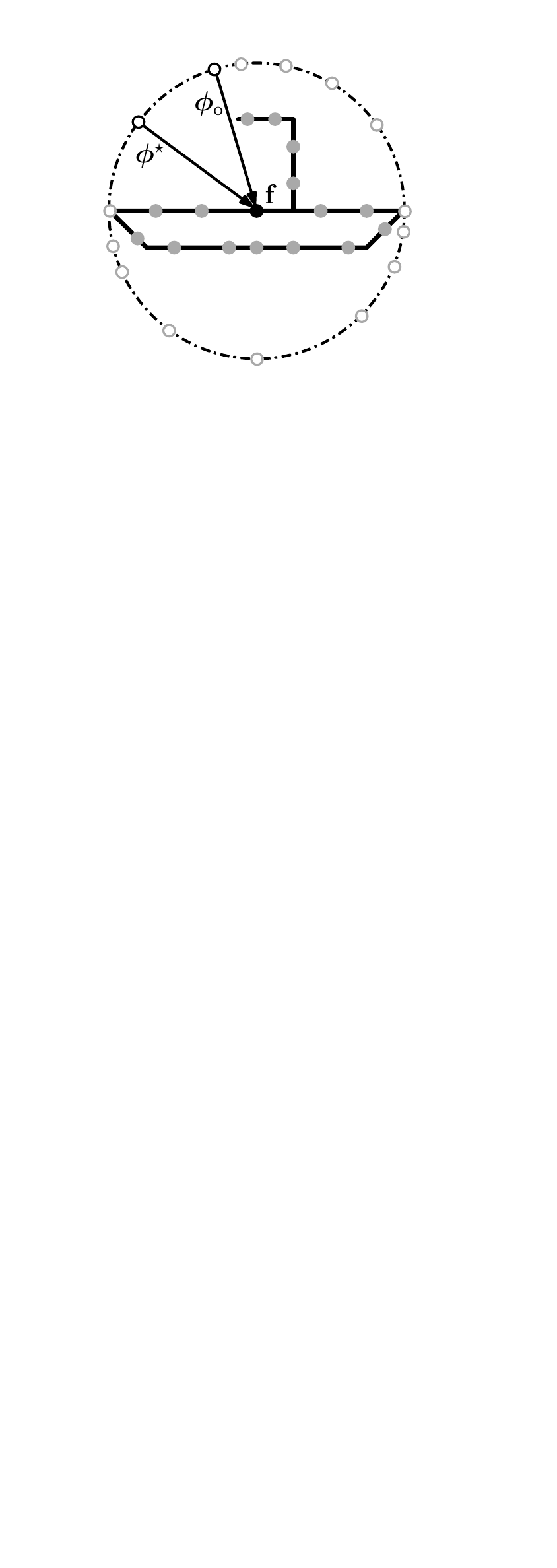}
  \caption{An illustration of the approach to propose unoccluded views. Points (grey dots) within a given radius of the frontier point (black dot), $\frontp$, are projected onto a sphere (grey circles) centered on the frontier. The optimal view orientation, $\viewo^\star$, is represented by a point on the sphere (black circle) which maximises the minimum distance to the projected points. The initial solution is the orientation of the view which first observed the frontier, $\viewo_\mathrm{o}$.}
  \figlabel{view_opt}
  \vspace{-2ex}
\end{figure}

The specific optimisation method depends on the distribution of the projected points. If they are spread over the full sphere then the smallest containing cap will be larger than a hemisphere and is found by minimising the distance of the plane from the sphere center,
\begin{argmini*}
    {\substack{\npoint\,\in\,\mathbb{R}^3,\,e\,\in\,[0,1]}}{e}
    {}{(\npoint^\star,e^\star):=}
    \addConstraint{e}{\leq \npoint^T\npoint}
    \addConstraint{e}{\geq \npoint^T\mathbf{q},\quad}{\mathbf{q} \in Q\,.}
\end{argmini*}
In this case the initial distance is one and the initial normal is the inverse of the observing view orientation, $\npoint = -\viewo_\mathrm{o}$. The minimax solution, $\optpoint$, is given by the inverse unit normal as this intersects the containing cap at the minimax point,
\[\optpoint = -\hat{\npoint}^\star\,,\]
where $\hat{\npoint}^\star$ denotes a unit vector in the direction of $\npoint^\star$.

If the projected points are contained in less than a hemisphere then the full sphere optimisation converges to a plane bisecting the sphere (i.e., $e^\star=0$). This indicates the smallest containing cap is smaller than a hemisphere. It can then be found by maximising the distance of the plane from the sphere center,
\begin{argmaxi*}
    {\substack{\npoint\,\in\,\mathbb{R}^3,\,e\,\in\,[0,1]}}{e}
    {}{(\npoint^\star,e^\star):=}
    \addConstraint{e}{\geq \npoint^T\npoint}
    \addConstraint{e}{\leq \npoint^T\mathbf{q},\quad}{\mathbf{q} \in Q\,.}
\end{argmaxi*}
In this case the initial distance is zero and the initial normal is the observing view orientation, $\npoint = \viewo_\mathrm{o}$. The minimax solution, $\optpoint$, is given by the unit normal as this intersects the containing cap at the minimax point, $\optpoint = \hat{\npoint}^\star$.

The maximin solution is the antipole of the minimax solution. It represents the direction of an unoccluded sight line starting at the frontier point and pointing towards free space. This means the orientation of the view proposed to observe the frontier along this line is equal to the minimax solution, $\viewo^\star = \optpoint$. The view position is then located at a viewing distance along the maximin solution.

Proactively handling occlusions by detecting occluding points and proposing unoccluded views of frontiers allows known occlusions to be avoided. This limits the use of incremental view adjustments to cases where the visibility of frontiers is obstructed by unknown occlusions, thereby requiring fewer views and less travelling to observe a scene. 

\subsection{Considering Scene Coverage}
\seclabel{view_sel}

The coverage of incompletely observed surfaces is improved by selecting a next best view to observe the greatest number of frontier points while moving the shortest distance. The visibility of frontier points from view proposals is evaluated with occlusion detection and captured in a frontier visibility graph. The next best view is selected by considering a set of view proposals close to the current sensor position and choosing the view from this set that can observe the most frontiers relative to the sensor travel distance \figref{view_sel}.

The frontier visibility graph is a directed graph, $\mathcal{G} = (M, E)$, that connects views with frontiers based on their visibility. Vertices in the graph, $\fvpair$, each represent an associated pair of a frontier point, $\frontp$, and its corresponding view proposal, $\viewf$. An edge, $(\fvpairm_j, \,\fvpairm_k) \in E$, denotes that the parent view, $\viewf_j$, can observe the child frontier point, $\frontp_k$.

The graph is updated after new sensor measurements are obtained and point classifications have been processed. Vertices representing points that are no longer frontiers are removed and new vertices are added to represent new frontier-view pairs. Updates to the graph connectivity (i.e., edges) are then computed for a subset of vertices defined by the visibility update limit, $\tau$. This constrains the computational cost of updating the graph by only evaluating a local region from which the next best view is likely to be chosen.

Connectivity is updated for vertices associated with the $\tau$-nearest view proposals to the current sensor position. All of the existing outgoing edges associated with these vertices are removed. New outgoing edges are then added from each vertex to any vertices whose associated view proposals are in the set of $\tau$-nearest views to the view proposal of the vertex and whose frontiers are visible from that view proposal.

A next best view, $\viewf_{i+1}$, is selected to observe the greatest number of frontier points while travelling the shortest distance from the current view. The frontier point associated with the vertex having the closest view proposal, $\fvpairm_c$, to the current view, $\viewidx{i}$, is required to be visible from the selected view. This is achieved by selecting the next best view from a vertex set, $M_c$, containing parent vertices of incoming edges to $\fvpairm_c$. Only view proposals that can observe more frontier points than $\fvpairm_c$ and have a greater number of outgoing vertex edges (i.e, outdegree) are considered, 
\[M_c := \{\mathbf{\fvpairm} \in M \,|\,(\fvpairm,\,\fvpairm_c) \in E \,\land\, \mathrm{deg}^+(\fvpairm) > \mathrm{deg}^+(\fvpairm_c) \}\,,\]
where 
\[\fvpairm_c = \argmin_{\fvpairm \in M}(||\viewp - \viewp_i||)\,,\]
and $\mathrm{deg}^+(\fvpairm)$ denotes the outdegree of a given vertex, $\fvpairm$.

The next best view is the view proposal with the greatest number of outgoing edges relative to the travel distance,
\[\fvpairm_{i+1} = \argmax_{\fvpairm \in M_c}\left(\frac{\mathrm{deg}^+(\fvpairm)}{||\viewp - \viewp_i||}\right)\,.\]
If none of the evaluated view proposals have a greater outdegree than $\fvpairm_c$ (i.e., $M_c \equiv \emptyset$) then the next best view is the closest view proposal, $\fvpairm_{i+1} = \fvpairm_c$.

Selecting a next best view with this approach ensures that the chosen view is both close to the current sensor position and has the best local coverage of insufficiently observed surfaces. This allows SEE++ to obtain complete scene observations using markedly fewer views and shorter travel distances than \gls{see} and the volumetric approaches. 

\begin{figure}[tpb]
 \centering
  \includegraphics[width=0.99\linewidth]{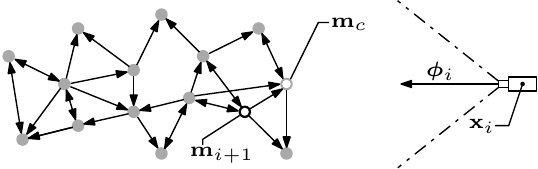}
  \caption{An illustration of the approach for selecting next best views that can observe the most frontier points while moving the least distance. Vertices (grey dots) in the frontier visibility graph are connected with edges denoting visibility (black arrows). The sensor represents the current view. The next best view is the view associated with the vertex (black circle), $\fvpairm_{i+1}$, that has the greatest outdegree relative to its distance from the sensor position, $\viewp_i$. It must also be able to observe the frontier point associated with the vertex (grey circle), $\fvpairm_c$, whose view is closest to the current sensor position.}
  \figlabel{view_sel}
  \vspace{-2ex}
\end{figure}

\begin{figure*}[tpb]
\centering
\captionsetup[subfigure]{}
\captionsetup[subfigure]{labelformat=empty}
\captionsetup[subfigure]{}
\subfloat[Newell Teapot ($1\,$m) \citem{Newell1975}]{\includegraphics[width=.24\linewidth]{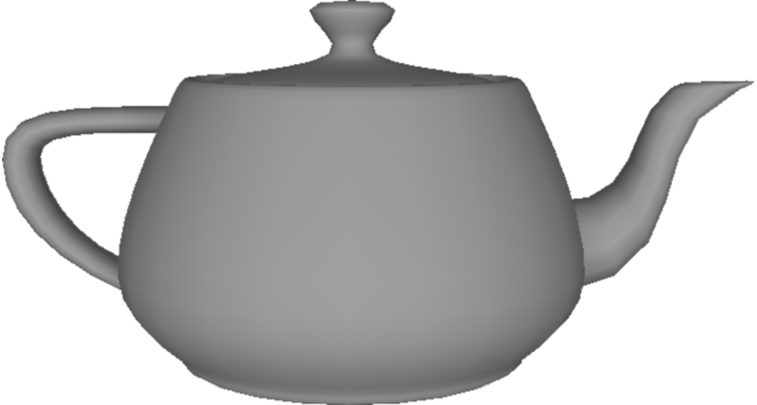}} \hfill
\captionsetup[subfigure]{labelformat=empty}
\subfloat[]{\includegraphics[width=.24\linewidth]{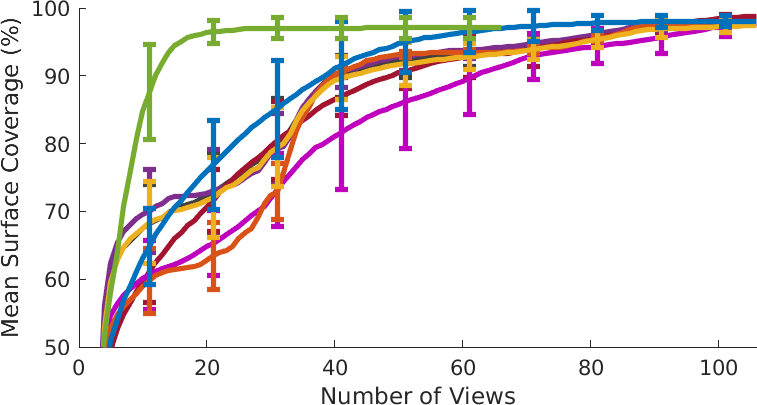}} \hfill
\subfloat[]{\includegraphics[width=.24\linewidth]{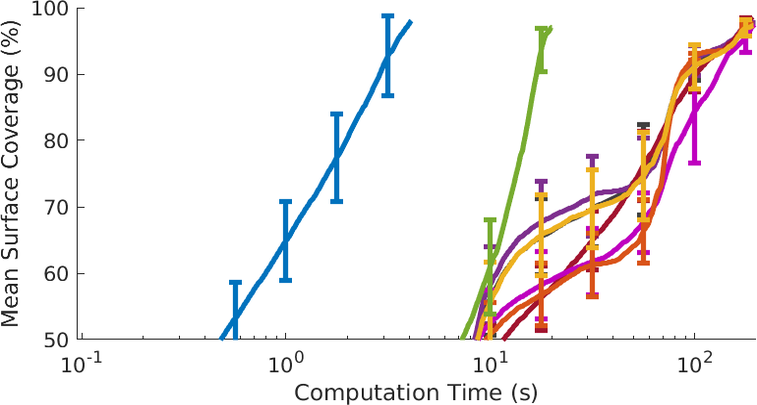}} \hfill
\subfloat[]{\includegraphics[width=.24\linewidth]{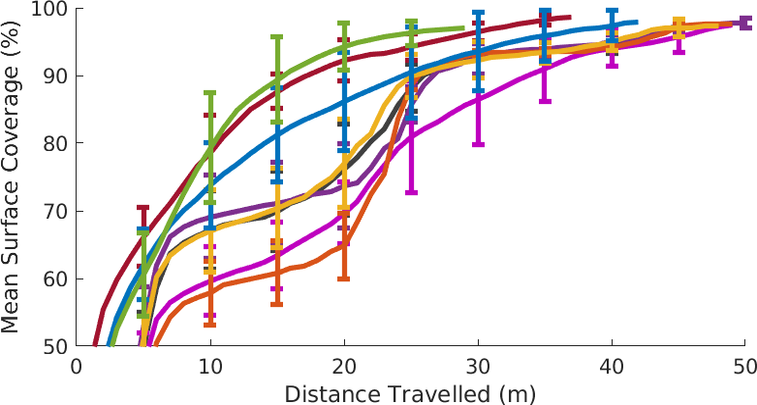}} \hfill
\captionsetup[subfigure]{}
\subfloat[Stanford Bunny ($1\,$m) \citem{Turk1994}]{\includegraphics[width=.24\linewidth]{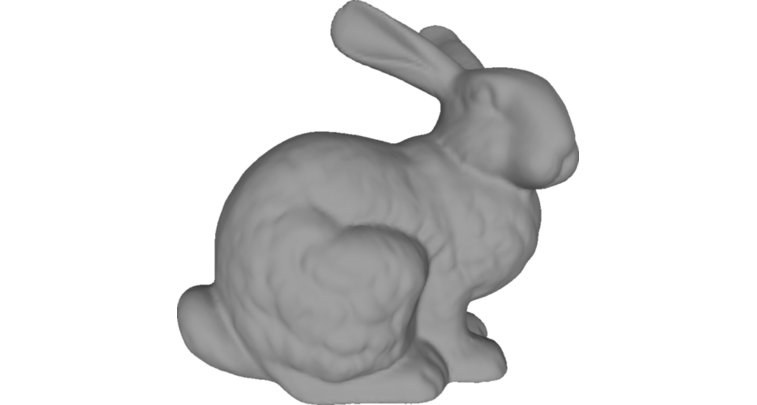}} \hfill
\captionsetup[subfigure]{labelformat=empty}
\subfloat[]{\includegraphics[width=.24\linewidth]{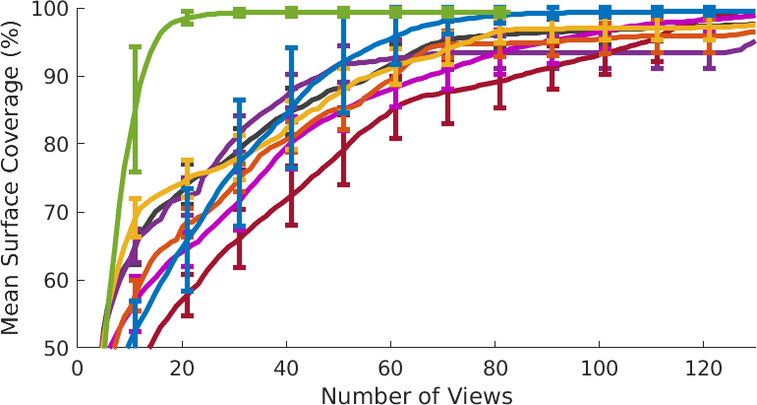}} \hfill
\subfloat[]{\includegraphics[width=.24\linewidth]{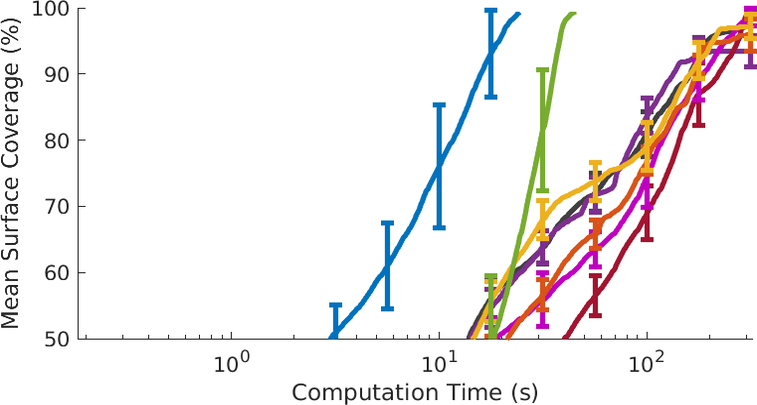}} \hfill
\subfloat[]{\includegraphics[width=.24\linewidth]{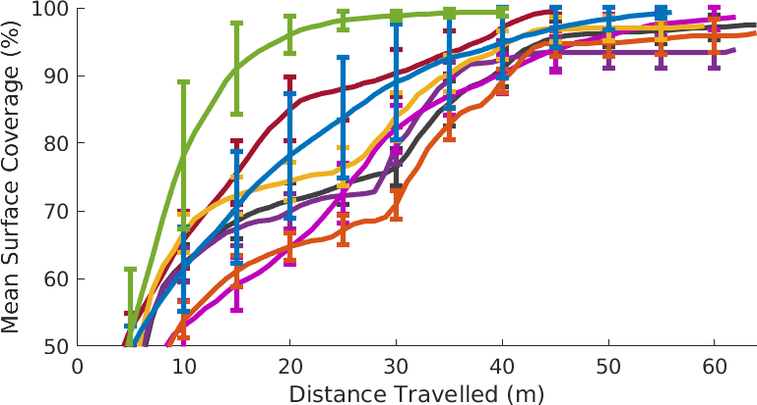}} \hfill
\captionsetup[subfigure]{}
\subfloat[Stanford Dragon ($1\,$m) \citem{Curless1996}]{\includegraphics[width=.24\linewidth]{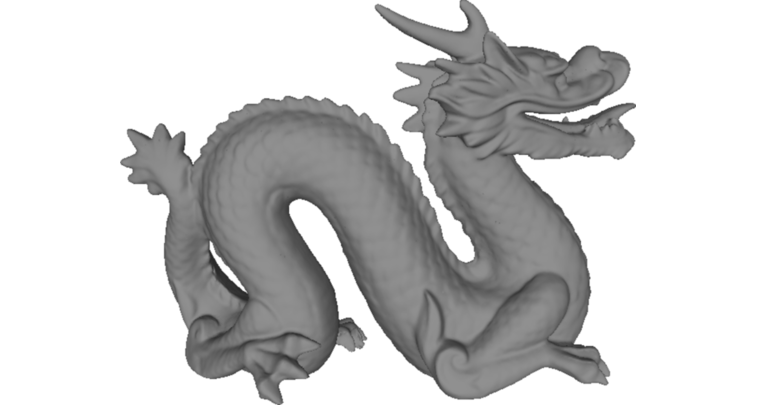}} \hfill
\captionsetup[subfigure]{labelformat=empty}
\subfloat[]{\includegraphics[width=.24\linewidth]{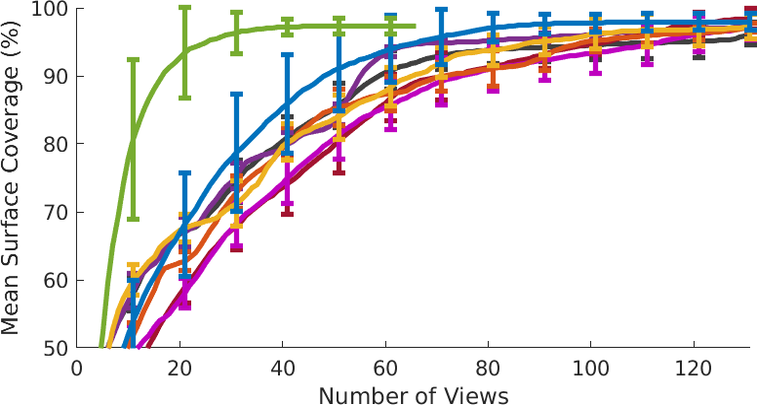}} \hfill
\subfloat[]{\includegraphics[width=.24\linewidth]{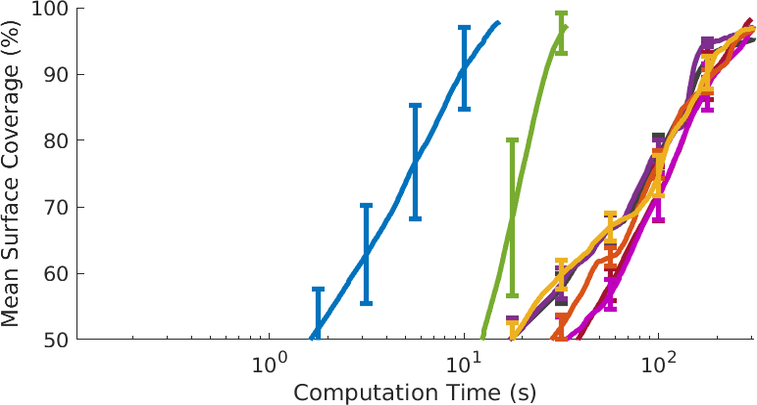}} \hfill
\subfloat[]{\includegraphics[width=.24\linewidth]{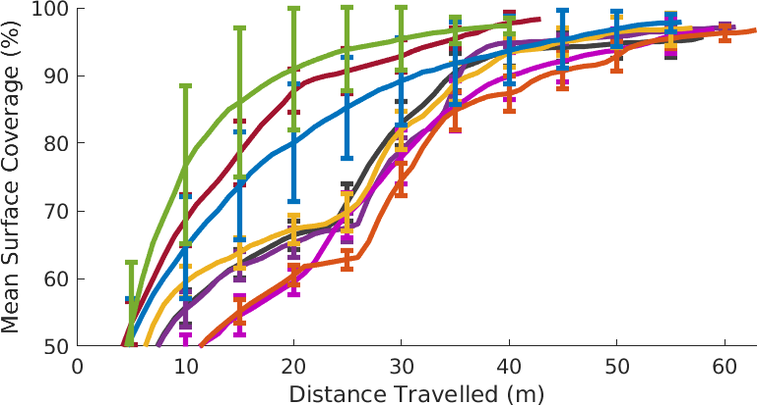}} \hfill
\captionsetup[subfigure]{}
\subfloat[Radcliffe Camera ($40\,$m) \citem{radcliffe}]{\includegraphics[width=.24\linewidth]{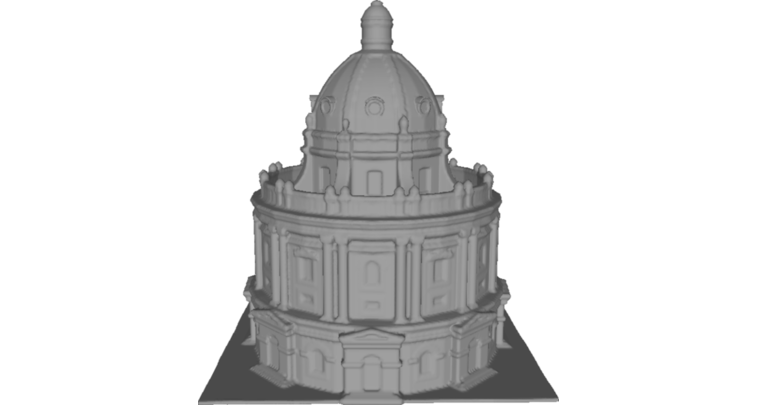}} \hfill
\captionsetup[subfigure]{labelformat=empty}
\subfloat[]{\includegraphics[width=.24\linewidth]{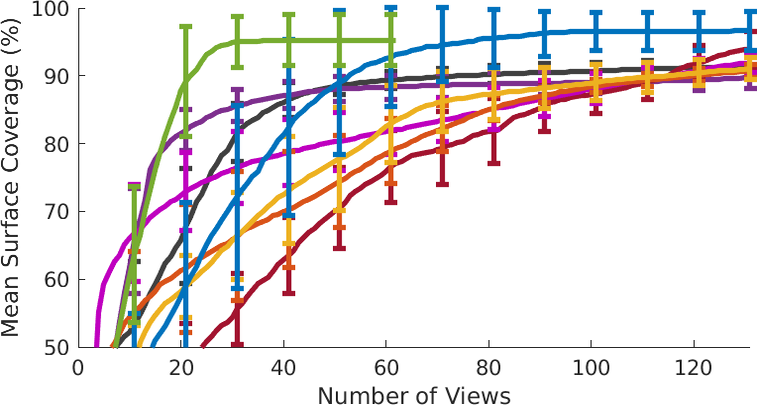}} \hfill
\subfloat[]{\includegraphics[width=.24\linewidth]{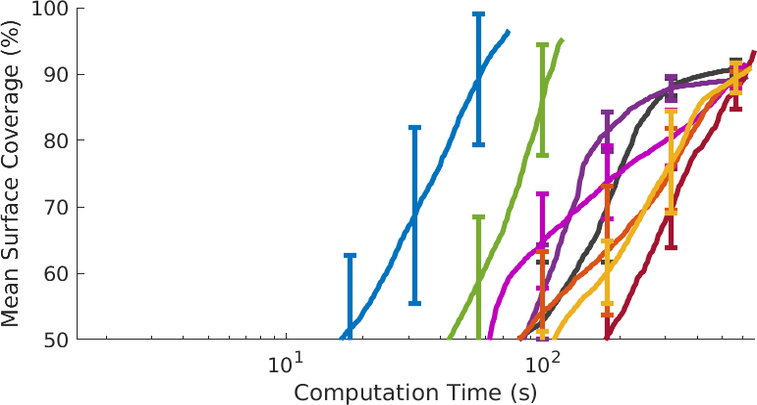}} \hfill
\subfloat[]{\includegraphics[width=.24\linewidth]{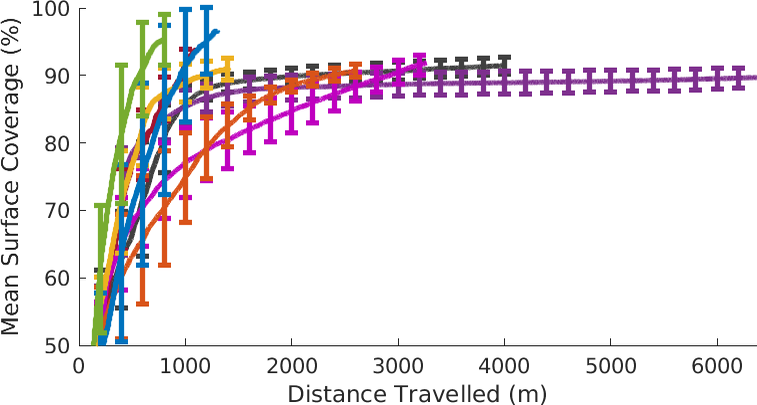}} \hfill
\captionsetup[subfigure]{}
\subfloat[Crocodile Skull  \citem{Schneider1801}]{\includegraphics[width=.24\linewidth]{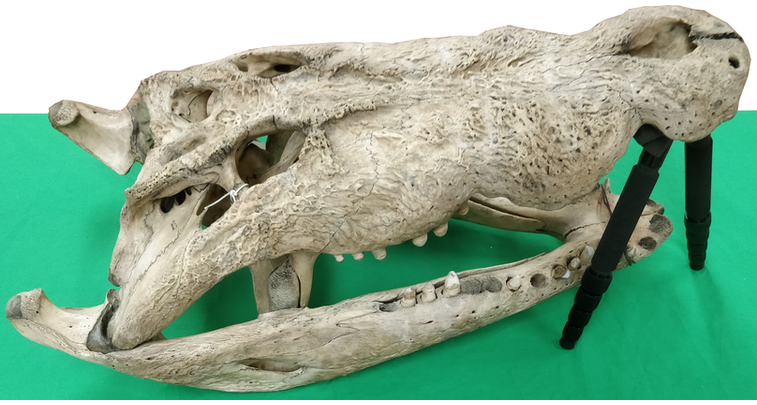}} \hfill
\captionsetup[subfigure]{labelformat=empty}
\subfloat[]{\includegraphics[width=.24\linewidth]{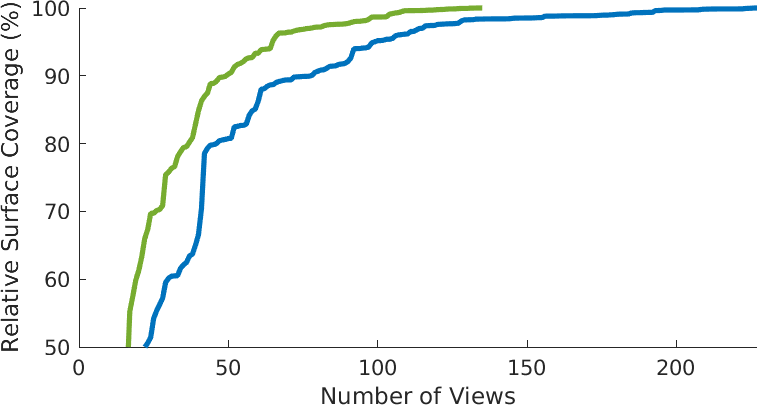}} \hfill
\subfloat[]{\includegraphics[width=.24\linewidth]{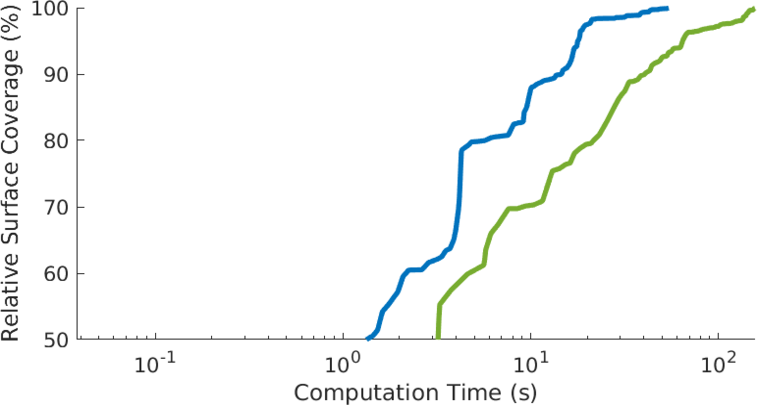}} \hfill
\subfloat[]{\includegraphics[width=.24\linewidth]{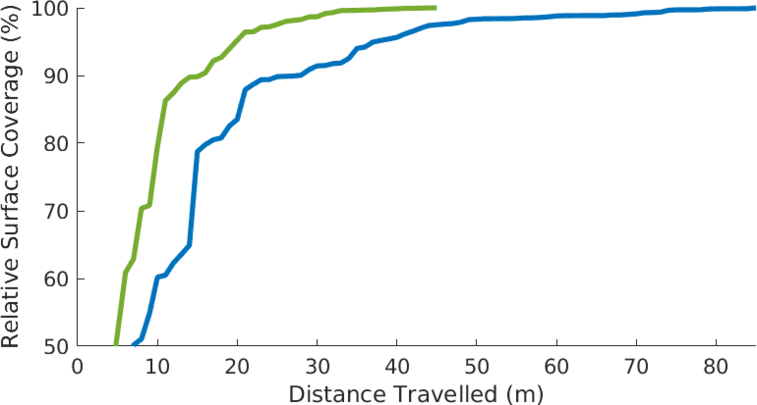}} \hfill \vspace{-2ex}
\subfloat[]{\includegraphics[width=.24\linewidth]{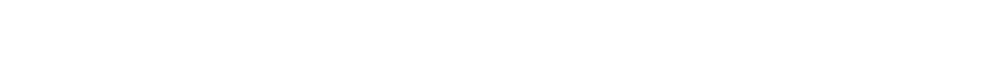}}
\subfloat[]{\includegraphics[width=.48\linewidth]{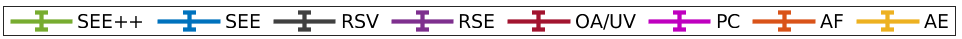}}
\caption{The top four rows show the performance of SEE++ versus \gls{see} \citem{Border2018} and state-of-the-art volumetric approaches \citem{Vasquez-Gomez2015, Kriegel2015, Delmerico2017} in a simulation environment on three standard models, (Newell Teapot \citem{Newell1975}, Stanford Bunny \citem{Turk1994} and Stanford Dragon \citem{Curless1996}), and a full-scale model of the Radcliffe Camera \citem{radcliffe}. The models used are presented in the left-most column. The graphs present the mean performance calculated from one hundred independent trials on each model. The mean surface coverage axes start at 50\% to improve the visual differentiation between the algorithm plots as they reach completion. Left to right they present the mean surface coverage vs the number of views, the mean overall planning time required and the mean distance travelled by the sensor. The error bars denote one standard deviation around the mean. The bottom row demonstrates the real-world performance of SEE++ versus \gls{see} \citem{Border2018} for the observation of a saltwater crocodile (\emph{Crocodylus porosus}) skull \citem{Schneider1801} using a hand-held Intel Realsense D435. A ground truth model of the crocodile skull was not available so the surface coverage metric for these experiments was computed relative to the final pointcloud observation obtained.}
\figlabel{results}
\end{figure*}
\begin{table*}[]
	\centering
	\resizebox{\textwidth}{!}{%
		\begin{tabular}{@{}lcccccccccccccccc@{}}
			\toprule
			\multicolumn{1}{c}{} & \multicolumn{4}{c}{Newell Teapot}                             & \multicolumn{4}{c}{Stanford Bunny}                            & \multicolumn{4}{c}{Stanford Dragon}                           & \multicolumn{4}{c}{Radcliffe Camera}                         \\ \cmidrule(lr){2-5} \cmidrule(lr){6-9} \cmidrule(lr){10-13} \cmidrule(l){14-17}
			& Views         & Coverage      & Time          & Distance      & Views         & Coverage      & Time          & Distance      & Views         & Coverage      & Time          & Distance      & Views         & Coverage      & Time          & Distance     \\ \midrule
			SEE++                & \textbf{22.3} & 97.1          & 20.1          & \textbf{29.7} & \textbf{31.5} & 99.4          & 46.0          & \textbf{40.9} & \textbf{35.5} & 97.3          & 34.0          & \textbf{40.6} & \textbf{27.1} & 95.3          & 121           & \textbf{806} \\
			SEE                  & 60.0          & 98.1          & \textbf{4.13} & 42.8          & 75.3          & 99.5          & \textbf{24.7} & 56.3          & 78.0          & 98.0          & \textbf{15.4} & 56.8          & 64.5          & \textbf{96.6} & \textbf{74.4} & 1302         \\
			RSV                  & 105           & 97.6          & 196           & 49.6          & 129           & 97.6          & 325           & 64.8          & 130           & 96.1          & 311           & 58.2          & 130           & 91.4          & 648           & 4008         \\
			RSE                  & 105           & 97.8          & 200           & 50.7          & 129           & 95.1          & 324           & 62.7          & 130           & 97.2          & 311           & 61.1          & 130           & 89.7          & 630           & 6375         \\
			OA/UV                & 105           & \textbf{98.8} & 194           & 37.8          & 129           & \textbf{99.5} & 313           & 45.1          & 130           & \textbf{98.4} & 300           & 43.1          & 130           & 93.9          & 675           & 1098         \\
			PC                   & 105           & 97.6          & 198           & 50.0          & 129           & 98.9          & 320           & 62.9          & 130           & 97.2          & 306           & 59.2          & 130           & 91.9          & 622           & 3272         \\
			AF                   & 105           & 97.6          & 196           & 49.9          & 129           & 96.4          & 326           & 64.3          & 130           & 97.1          & 306           & 63.8          & 130           & 90.7          & 628           & 2615         \\
			AE                   & 105           & 97.4          & 196           & 48.5          & 129           & 97.4          & 324           & 59.8          & 130           & 97.3          & 311           & 58.0          & 130           & 91.0          & 650           & 1409         \\ \bottomrule
		\end{tabular}%
	}
	\begin{tablenotes}[flushleft]
	\item Table 1.\quad The mean number of views captured, the mean surface coverage obtained, the mean computation time used and the mean travel distance required to observe three one-metre standard models (Newell Teapot \citem{Newell1975}, Stanford Bunny \citem{Turk1994} and Stanford Dragon \citem{Curless1996}) and a $40$ metre model of the Radcliffe Camera \citem{radcliffe}, calculated from $100$ experiments with SEE++, \gls{see} and state-of-the-art volumetric approaches \citem{Vasquez-Gomez2015, Kriegel2015, Delmerico2017}. The best performance values are bolded. Note that SEE++ obtains equivalent surface coverage using significantly fewer views and less travel distance than all of the other evaluated approaches.
	\end{tablenotes}
	\tbllabel{tables}
\end{table*}

\section{Evaluation}
\seclabel{evaluation}

SEE++ is compared to \gls{see} \citem{Border2018} and state-of-the-art volumetric approaches, (AF, \citem{Vasquez-Gomez2015}; AE, \citem{Kriegel2015}; and RSV, RSE, OA, UV, PC, \citem{Delmerico2017}), in a simulation environment on three standard models, (Newell Teapot \citem{Newell1975}, Stanford Bunny \citem{Turk1994} and Stanford Dragon \citem{Curless1996}), and a full-scale model of the Radcliffe Camera \citem{radcliffe}. The implementations of the volumetric approaches used to produce the presented results are provided by \citem{Delmerico2017}. 

These experimental results also correct a mistake in \citem{Border2018}. Those previous results had erroneously used a nonuniform distribution of view proposals for the volumetric approaches.

Real-world observations of a saltwater crocodile (\emph{Crocodylus porosus}) skull \citem{Schneider1801} using SEE/SEE++ are also presented.

\subsection{Sensors}
The simulation experiments are performed using virtual sensors defined by a field-of-view, $\theta_x$ and $\theta_y$, and resolution, $\omega_x$ and $\omega_y$. The standard models are observed using a simulated Intel Realsense D435 ($\theta_x = 69.4\degree$, $\theta_y = 42.5\degree$, $\omega_x = 848$~px and $\omega_y = 480$~px). The Radcliffe Camera is observed with a high-resolution sensor ($\theta_x = 60\degree$, $\theta_y = 40\degree$, $\omega_x = 2400$~px and $\omega_y = 1750$~px). Measurements are obtained by raycasting the surface mesh of a model with the virtual sensor. Sensor noise is simulated by adding Gaussian noise ($\mu = 0\,$m, and $\sigma = 0.01\,$m) to the observed points.  

The real-world experiments are performed with a hand-held Intel Realsense D435 ($\theta_x = 69.4\degree$, $\theta_y = 42.5\degree$, $\omega_x = 848$~px and $\omega_y = 480$~px). The sensor pose is obtained using a Vicon system to enable the hand-held alignment of views.

\subsection{Parameters}

The standard model simulation experiments use a desired density of $\rho = 146000$~points per m$^3$ with a resolution of $r = 0.017\,$m. The Radcliffe Camera simulation experiments use a desired density of $\rho = 213$~points per m$^3$ with a resolution of $r = 0.15\,$m. The crocodile skull real-world experiments use a desired density of $\rho = 10^6$~points per m$^3$ with a resolution of $r = 0.01\,$m. In all of the experiments SEE++ uses an occlusion search distance of $\psi = 1\,$m and a visibility update limit of $\tau = 100$~views.

The voxel grid resolution used by the volumetric approaches in the simulation experiments is equal to the SEE/SEE++ resolution parameter, $r$, used for each model. 

The view distance in all experiments is set such that the density of sensor measurements in the viewing frustum is equal to the desired measurement density, i.e.,
\[d = {\left(\frac{3\omega_x\omega_y}{4\rho\tan{0.5\theta_x}\tan{0.5\theta_y}}\right)}^\frac{1}{3}\,.\]

The simulation experiments are run one hundred times per algorithm on each model. SEE/SEE++ are run until their completion criteria is satisfied. The view limit for the volumetric approaches on each model is set to the maximum number of views \gls{see} required to complete an observation. 

Potential views for the volumetric approaches are sampled from a view sphere surrounding the scene as in \citem{Vasquez-Gomez2015, Delmerico2017}. Kriegel et al. \citem{Kriegel2015} does not restrict views to a view surface but we use the implementation provided by \citem{Delmerico2017} which does. The radius of the view sphere is set to the sum of the view distance and the mean distance of points in the model from their centroid. The number of views sampled on the view sphere is defined as $2.4$ times the view limit, as in \citem{Delmerico2017}.

In all experiments, a minimum distance, $\epsilon$, between sensor measurements is enforced to maintain an upper bound on memory consumption and computational cost. This distance is set based on the desired density, $\epsilon = \sqrt{\rho^{-1}}$. New measurements are only added to an observation if their $\epsilon$-radius neighbourhood contains no existing points.

\subsection{Metrics}
The algorithms are evaluated using surface coverage, computational time and sensor travel distance as defined in \citem{Border2018}. The registration distance used to compute surface coverage for the simulation experiments is $r_\mathrm{d} = 0.005\,$m for the standard models and $r_\mathrm{d} = 0.05\,$m for the Radcliffe Camera.

A ground truth model of the crocodile skull was not available so the surface coverage metric for these experiments was computed relative to the final pointcloud observation obtained. The registration distance used is $r_\mathrm{d} = 0.005\,$m. 

\section{Discussion}

\begin{figure}[tpb]
	\centering
	\captionsetup[subfigure]{labelformat=empty}
	\subfloat[]{\includegraphics[width=0.85\linewidth]{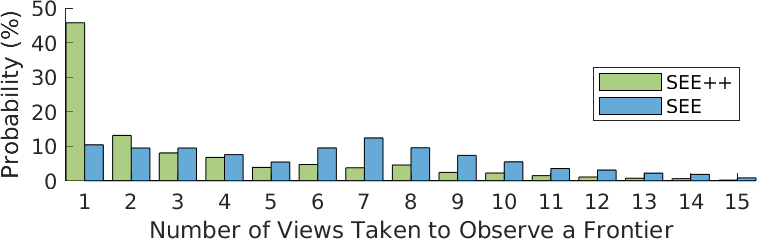}} \vspace{-2ex}
	\subfloat[]{\includegraphics[width=0.85\linewidth]{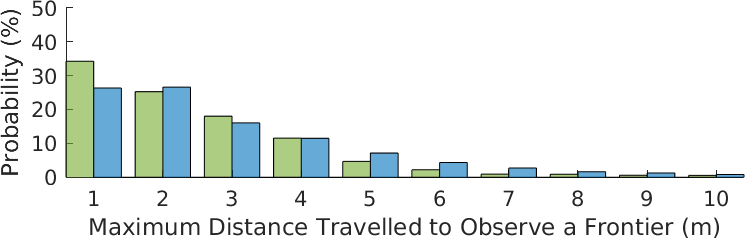}}
	\caption{A statistical analysis of the view proposal and selection performance of \gls{see} and SEE++ calculated from the experiments on the standard models \secref{evaluation}. SEE++ is $4.5$ times more likely to observe a frontier with a single view than \gls{see} (top graph) and travels less than $3\,$m to observe a frontier point $77\%$ of the time while \gls{see} only observes a frontier within this distance $69\%$ of the time (bottom graph).}
	\figlabel{discussion}
\end{figure}

\begin{table}[tpb]
	\begin{adjustbox}{center}
		\begin{tabular}{@{}lcc@{}}
			\toprule
			\multicolumn{1}{c}{} & \begin{tabular}[c]{@{}c@{}}Frontiers\\ Observed\end{tabular} & \begin{tabular}[c]{@{}c@{}}Surface\\ Coverage (\%)\end{tabular} \\ \midrule
			SEE & 4.49 & 1.39 \\
			SEE++ & \textbf{6.51} & \textbf{3.29} \\ \bottomrule
		\end{tabular}
	\end{adjustbox}
	\begin{tablenotes}[flushleft]
		\item Table 2.\quad The mean number of frontiers observed and surface coverage obtained per view for \gls{see} and SEE++ calculated from the one hundred experiments on each of the standard models.
	\end{tablenotes}
	\tbllabel{view_result}
	\vspace{-2ex}
\end{table}

The experimental results (\tfigref{results}; Table 1) show that SEE++ consistently outperforms \gls{see} and the evaluated state-of-the-art volumetric approaches by requiring significantly fewer views and shorter travel distances to obtain an equivalent quality of observations. This performance improvement is achieved while maintaining a considerably lower computation time than the volumetric approaches. 

These results demonstrate the value of proactively handling occlusions and considering scene coverage in unstructured representations for NBV planning. SEE++ is more efficient as the proposed views obtain greater scene coverage and are more successful at observing their target frontier points (\tfigref{discussion}; Table 2). The overall computational times of SEE++ are still significantly lower than the evaluated volumetric approaches despite the relative per-view increase compared to \gls{see}. This is because proactively accounting for the scene structure when planning next best views significantly reduces the number of views required for a complete observation.

The independent contributions of the methods for proactively considering occlusions and scene coverage are evidenced by a statistical analysis of the distance travelled, number of frontiers observed and surface coverage obtained. These metrics are calculated per frontier point \figref{discussion} and per view (Table 2) from the standard model experiments. A complete investigation of the performance improvements achieved by independently including the proactive occlusion handling and frontier visibility graph is presented in \citem{BorderThesis}.    

Accounting for known occlusions when proposing views increases the likelihood that a frontier point will be visible and decreases the number of view adjustments required to observe it. This improves the efficiency of frontier observations by reducing the number of views and travel distance required per frontier \figref{discussion}. SEE++ is $4.5$ times more likely to observe a frontier point with a single view than \gls{see}. The distance travelled by SEE++ to observe a frontier point is less than $3\,$m in $77\%$ of cases while \gls{see} only observes a frontier within the same distance $69\%$ of the time. 

Selecting next best views which observe the most frontier points while travelling short distances increases the number of frontiers observed and surface coverage obtained per view (Table 2). SEE++ observes $45\%$ more frontier points and obtains $137\%$ greater surface coverage per view than \gls{see}. This allows SEE++ to capture significantly fewer views than \gls{see} while obtaining equivalent scene observations.

\section{Conclusion}  

This paper presents proactive methods for handling occlusions and considering scene coverage with a \gls{nbv} planning approach that uses an unstructured representation. The occlusion handling technique detects occluded views and applies an optimisation strategy to propose alternative unoccluded views. The frontier visibility graph encodes knowledge of which frontiers are visible from proposed views and is used to select next best views that most improve scene coverage.

The value of these presented techniques is demonstrated by extending \gls{see} to create SEE++. Proactively accounting for known occlusions when proposing views increases the likelihood that target frontier points will be successfully observed without requiring incremental view adjustments. Assessing the visibility of frontier points from proposed views when selecting a next best view improves the scene coverage attained from each view while retaining relatively short travel distances between views. A significant improvement in observation efficiency is achieved by integrating these methods with an unstructured scene representation.

Experimental results demonstrate that SEE++ outperforms \gls{see} and the evaluated volumetric approaches by requiring fewer views and less travelling to obtain an equivalent quality of observations. SEE++ uses greater computation times than \gls{see} but retains lower times than the volumetric approaches.

We plan to use SEE++ for observing small indoor scenes using an RGB-D camera affixed to a robotic arm and larger outdoor scenes with a LiDAR sensor mounted on an aerial platform. Information on an open-source version of SEE++ is available at \url{https://robotic-esp.com/code/see}.

\renewcommand*{\bibfont}{\footnotesize}
{\renewcommand{\markboth}[2]{}
\renewcommand*{\UrlFont}{\rmfamily}
 \printbibliography}

\end{document}